\documentclass[11pt,a4paper]{article}
\usepackage{abstract}
\usepackage{url}
\usepackage{graphicx}
\usepackage{epstopdf}
\usepackage{upquote}
\usepackage{textcomp}

\usepackage{newtxtext,newtxmath}

\usepackage{fontspec}
\usepackage[bidi=default,english]{babel}
\babelprovide[import]{arabic}

\newfontfamily\arabicfont[
  Path=./,
  UprightFont=Amiri-Regular.ttf,
  Script=Arabic
]{Amiri}
\newcommand{\ar}[1]{\foreignlanguage{arabic}{\arabicfont #1}}
\usepackage{natbib}
\usepackage[acceptedWithA]{tacl2021v1}

\usepackage{xspace,mfirstuc,tabulary}

\newif\iftaclinstructions
\taclinstructionsfalse 
\iftaclinstructions
\renewcommand{\confidential}{}
\renewcommand{\anonsubtext}{(No author info supplied here, for consistency with
TACL-submission anonymization requirements)}
\newcommand{\instr}
\fi

\iftaclpubformat 

\else

\fi

\usepackage{amsmath}

\usepackage{subcaption} 
\usepackage{latexsym}
\usepackage{color,soul}

\usepackage{microtype}

\usepackage{inconsolata}

\usepackage{graphicx}

\usepackage{subcaption}
\newcommand{\say}[1]{``#1''}
\usepackage{tablefootnote}

\usepackage{tabularx}
\usepackage{booktabs}
\usepackage{array}

\title{Same Lesson, Different Story: Cross-Lingual Reconstruction of Cultural Narratives in Large Language Models}

\author{
Jory Alshaalan,
Haya Albaker,
Abeer Aldayel,
Aljawharah Alabdullatif,
Rehab Alahmadi \\
College of Computer and Information Sciences \\
King Saud University, Saudi Arabia \\
\texttt{\{jooryabdullahsh,haya.aalbaker\}@gmail.com} \\
\texttt{\{aabeer,aabdullatif1,ralahmadi\}@ksu.edu.sa}
}

\begin{document}
\maketitle
\begin{abstract}
The evaluation of cultural grounding context becomes complex when multiple cultures convey the same moral lesson. This challenge is particularly relevant to large language models (LLMs), which produce narratives across a wide range of languages and cultural contexts. However, it remains uncertain whether these models preserve culturally grounded meaning when equivalent moral lessons are conveyed through distinct cultural forms. This study introduces a multilingual evaluation narrative framework that integrates a cross-linguistic collection of 414 proverbs spanning 15 languages and uses four LLMs to generate 13k narratives. By employing semantically equivalent proverbs as culturally grounded prompts, the analysis assesses whether models preserve meaning across languages, how cross-lingual conditioning influences narrative realization, and whether different model families converge on similar interpretations. Results indicate that cross-lingual prompting largely preserves proverb-level semantic meaning while systematically redistributing agency, social positioning, and narrative structure. Additionally, strong inter-model convergence is observed in both monolingual and cross-lingual settings, suggesting that multilingual LLMs rely on shared semantic abstractions despite architectural and linguistic differences. These findings shed light on the need for more comprehensive evaluations of cultural grounding. Relying exclusively on semantic similarity in multilingual narrative assessments may overestimate cultural preservation by neglecting culturally meaningful variations in narrative expression.
\end{abstract}

\section{Introduction}
\begin{figure*}[t!]
    \centering
    \includegraphics[width=\textwidth]{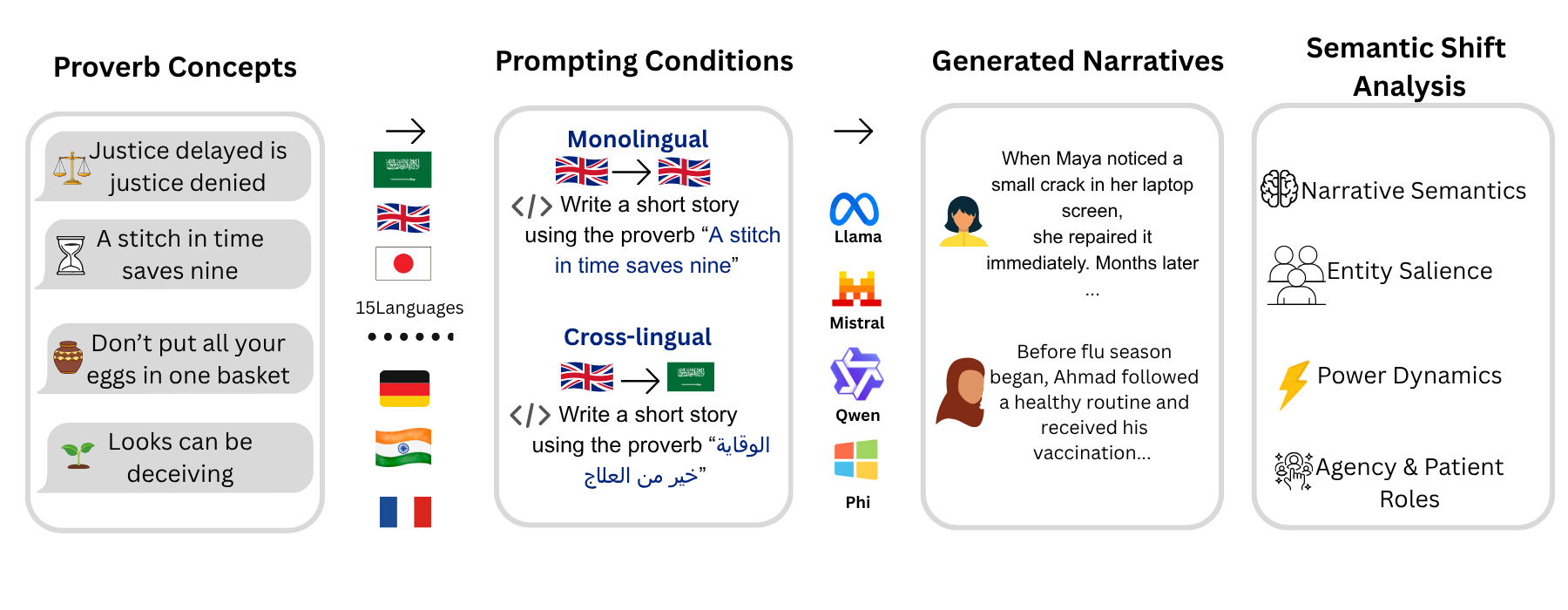}
    \caption{
    Overview of the proverb narrative generation and semantic shift analysis pipeline. 
    Starting from proverb concepts, we evaluate multilingual realizations across 15 languages and two prompting conditions: monolingual, where the proverb and prompt are in the same language, and cross-lingual, where the proverb is prompted through another language. 
    Four LLMs generate short narratives, which are then analyzed along four dimensions: narrative semantics, entity salience, power dynamics, and agency/patient roles.
    }
    \label{fig:proverbpipeline}
\end{figure*}

Stories reflect cultural interpretations of moral principles and social expectations, shaping how values are passed down through generations. Cross-cultural analyses of human-authored folktales have shown systematic regional variation in how moral values and social roles are expressed across societies \citep{Wu2023-xq}. Similar regional variation has been observed even in computationally generated narratives, where cross-cultural analyses have revealed systematic regional variation in narrative moral content and character agency \citep{Mitran2025-ue,Toro-Isaza2023-wd}. 
Although many social values and norms appear in different societies, the way they are told depends on the cultural and language context, leading to unique storytelling styles and emphases. This mix of common moral ideas and culturally specific storytelling makes narratives an important lens for studying cross-cultural understanding and how it can be computationally modeled.

Narrative generation has become a well-known capability of large language models (LLMs)~\cite{Brown2020-qt,Teleki2025-nu}, yet our understanding of how these models construct stories across languages, cultures, and narrative traditions remains limited~\citep{Adilazuarda2025-ii}. Most evaluations of generative models focus on surface-level quality or task-specific correctness, overlooking deeper narrative properties such as thematic development, character structure, stylistic choices, and coherence~\cite{Pranida2025-so,Rashkin2018-do}. These aspects are central to narrative reasoning and cultural grounding, particularly when models are prompted with culturally rooted material such as folklore, mythology, and proverbs, where cultural nuance, implicit knowledge, and narrative conventions play a meaningful role. At the same time, culturally rooted storytelling presents a unique challenge for LLMs: the challenge of generating narratives that are structurally coherent, thematically meaningful, and contextually appropriate across languages. Previous work has focused on the factual correctness of generated stories \citep{Bhagat2026-uu}. Consistent with~\citep{Zhou2025-di}, cultural meaning is not reducible to static facts or survey-style knowledge. Instead, it is constructed through contextualized narratives and lived experiences that differ across communities. In this work, the discussion moves from “Do models know culture?” to: “At what representational level is cultural meaning encoded?”. Little research has investigated how LLMs generate full narratives from culturally grounded prompts, and even less is known about how such narratives vary across languages and model families. More specifically, prior work has largely examined how language models express a particular moral through different narrative patterns or whether models retain cultural knowledge in isolation. In contrast, we focus on a complementary question: \textit{What happens when the same moral meaning is carried across languages?} We use proverbs as a proxy for story generation because they encode shared moral wisdom across different cultures, while literal articulation can vary, reflecting the base culture norms~\citep{Liu2024-ab}.

 In particular, the study addresses the following research questions:

\noindent\textbf{RQ1: Semantic Stability and Transformation under Cross-Lingual Conditioning.}
To what extent do LLMs preserve semantic meaning across monolingual and cross-lingual settings, and how does cross-lingual conditioning redistribute semantic roles within narratives?

\vspace{0.5em}

\noindent\textbf{RQ2: Inter-Model Semantic Convergence.}
To what extent do different LLMs produce similar narratives under both monolingual and cross-lingual conditions?

\vspace{0.5em}

\noindent\textbf{RQ3: Level of Cultural Representation.}
At what representational level is cultural meaning encoded: abstract semantic lessons or language-specific narrative realizations?

Our study contributes a multilingual evaluation framework (Figure~\ref{fig:proverbpipeline}) that combines a cross-language proverb collection with a large corpus of LLM-generated narratives, enabling controlled analysis of challenging case of parallel moral meaning through culturally grounded story generation across 15 languages and four model families. Through this framework, we show that cross-lingual prompting largely preserves proverb-level semantic meaning while systematically redistributing entities, agency, and social positioning, revealing a distinction between semantic preservation and narrative realization. We further demonstrate that strong inter-model convergence persists even under culturally grounded multilingual prompting, extending recent observations of model homogeneity to cross-lingual narrative generation. 

\section{Related work}
The growing use of large language models across language and cultural contexts has prompted increasing interest in evaluating their cultural competence. Much of this work has focused on testing whether models possess discrete cultural knowledge, such as facts about customs, foods, or social norms, rather than examining how cultural meaning is constructed and preserved through narrative generation \cite{Adilazuarda2025-ii}. Research using proverbs as a lens for cultural reasoning has shown that while multilingual LLMs possess knowledge of proverbs to varying degrees, this knowledge does not guarantee culturally appropriate reasoning, and significant gaps emerge when models reason with proverbs translated from another language \citep{Liu2024-ab}. Narrative is a particularly revealing context for extending this question further. Stories require models to make choices about structure, agency, and framing, and these are precisely the dimensions along which cultures differ in how they encode and transmit meaning. A number of studies~\citep{Zhou2024-iz,Swanson2017-ip} have therefore examined story generation within single-language, topic domain or single-culture settings, revealing consistent gaps between model outputs and culturally authentic narratives.

\citet{Moosavi-Monazzah2025-zl} introduce PerCul, a story-driven benchmark that evaluates LLMs on Persian cultural knowledge through narratives that implicitly reference culturally specific concepts. Their evaluation of multiple state-of-the-art models reveals that models consistently rely on surface-level narrative details rather than deeper cultural inference when identifying culturally embedded concepts. Critically, translating the benchmark into English results show a drop in model performance across the best-performing model families which demonstrate that cultural knowledge in these models does not transfer across languages and that language plays a meaningful role in how cultural meaning is accessed and expressed. For Javanese and Sundanese, \citet{Pranida2025-so} construct a culturally grounded story comprehension benchmark for these languages, using stories that embed local names, foods, customs, and rituals as cultural cues, authored and verified by native speakers. Their results show that models covering the target languages produce culturally inappropriate narrative endings even when cultural content is explicitly present in the prompts and stories, indicating that linguistic coverage does not guarantee cultural grounding in narrative generation. Another work by~\citep{Hobson2024-jp}, examine the story generation using different domain of data covering TV, books, Folktales, Reddit 
stories and show that same moral value can be represented through multiple narrative.

A separate line of work has examined the diversity of LLM outputs on open-ended generative tasks. \citet{Jiang2025-cr} document what they call the Artificial Hivemind effect, finding that despite differences in architecture and training, LLMs produce strikingly homogeneous outputs when given open-ended prompts, both within a single model and across model families. More recently, a study by~\cite{Fundal2026-fj} on human–LLM co-writing examines the narrative agency that derive the story. In their study they explicitly model narrative agency and semantic novelty rather than treating stories as bags of words 

 Overall, the previous work lacks a proper examination of whether a story's cultural meaning embedded in a prompt survives when generation crosses language boundaries, and specifically, how cross-lingual conditioning affects the distribution of semantic roles within narratives, especially when a parallel moral meaning is carried between languages and cultures. More specifically, previous work has generally emphasized the distinction among narrative morals, character agency, and narrative realization. Our study extends this perspective to multilingual generation via cross-lingual permutation to examine the stability of proverb-level moral content and to test variation in entities and social roles. 

\section{Dataset}
We use proverbs as culturally grounded prompts to examine the capability of story generation. The proverbs are a strong representation of folk wisdom and social norms~\citep{Syzdykov2014-wd}. As shown in a recent study, the proverb has been used as a proxy for evaluating cultural reasoning in LLMs to examine the challenging nature of the parallel meanings of proverbs across cultures~\citep{Liu2024-ab}. These folk proverbs embody various aspects of social values and cultural norms. For instance, \say{\ar{اللي مايعرف الصقر يشويه}}, which in literal translation means \say{He who does not know a falcon will roast it}. This proverb has an equivalent proverb in English: \say{You never miss the water till the well runs dry}, with a similar moral about showing appreciation for something valuable before it is lost. Yet the Arabic version used a symbol representing the Arab cultural value \say{Falcon}, a figure closely associated with prestige and social status in Arab cultures. On the other hand, in the English version, it is figuratively represented as \say{water}. Therefore, the generated stories are not treated as culture in its entirety, but as a proverb-grounded cultural narrative. 
\paragraph{Cross-Language Proverb.}
To evaluate models’ ability to generate culturally aligned narratives across languages, we constructed a parallel-proverb dataset from a curated list of widely used proverbs. We collect proverbs from the Tatoeba database\footnote{\url{https://tatoeba.org}} using the \texttt{proverb} tag and retrieve all available cross-language translations through the Tatoeba API. The dataset includes English and non-English variants covering multiple languages, along with explanations of cultural context. Using this dataset, we prompted multiple LLMs to generate short stories based on each proverb in both languages.\footnote{The dataset will be released upon publication}.

\paragraph{The Evaluation Corpus.} As shown in Table~\ref{tab:model_breakdown}, the data comprises 414 proverb concepts spanning 15 languages: Arabic, English, French, Spanish, German, Russian, Japanese, Portuguese, Italian, Polish, Turkish, Persian, Dutch, Ukrainian, and Hindi. Cross-lingual prompts are constructed by translating proverb meanings across this language set, with English and Arabic serving as source proverb languages. For each proverb, we generate narratives under two conditions: (i) a monolingual setting in which the model receives the original proverb, and (ii) a cross-lingual setting in which the model is prompted with a translated counterpart. We collect outputs from four language models (Llama-3, Mistral, Qwen, and Phi-3). The Arabic source data contains 111 unique proverb instances with complete multilingual translations, whereas the English data contains 1,545 unique instances, resulting in an asymmetric but naturally occurring cross-lingual evaluation set. Overall, the resulting corpus is  6,624 monolingual narratives and 6,346 cross-lingual narratives. Following model-specific alignment of monolingual and cross-lingual generations, the final dataset used for the analyses consists of 6,346 paired narrative comparisons.

\begin{table}[t]
\centering
\small
\setlength{\tabcolsep}{4pt}
\renewcommand{\arraystretch}{1.1}
\begin{tabular}{lrr}
\toprule
\textbf{Model} & \textbf{Mono} & \textbf{Cross} \\
\midrule
Llama-3 & 1,656 & 1,656 \\
Mistral & 1,656 & 1,656 \\
Qwen & 1,656 & 1,378 \\
Phi-3 & 1,656 & 1,656 \\
\midrule
\textbf{Total} & \textbf{6,624} & \textbf{6,346} \\
\bottomrule
\end{tabular}
\caption{Narrative generations by model and prompting condition. Qwen produced 278 fewer valid cross-lingual narratives than expected due to incomplete outputs and formatting artifacts. These instances were removed during data cleaning prior to analysis.}
\label{tab:model_breakdown}
\end{table}

\section{Methodology}
Narratives demonstrate social meaning through participants' expressions and their relations to events~\citep{Laszlo2010-wf}.
Drawing on Social Cognitive Theory~\citep{Bandura1986-hm}, which declares that agency can influence events and outcomes, therefore we further examine the positioning of entities. This experimental practice has been used in prior work on narrative analysis, in which actions performed on/upon entities are studied to reveal social meaning~\citep{Bamman2013-kj}. Moreover, Sociocultural theories assert that cultural meaning emerges through processes of social positioning rather than through static facts~\citep{Zhou2025-di}.
Thus, to evaluate narrative competence and cultural grounding in large language models, we apply a multi-layered computational narrative analysis pipeline that includes these two complementary dimensions. We focused on interpreting the semantics of the narrative generated in cross-lingual and monolingual settings using methods that gauge (1) the homogeneity of the generated narrative and (2) the social dynamics of the entity's power. This methodology quantifies the semantic and thematic aspects of each generated story, enabling clear comparisons across languages and model families.
\subsection{Semantic Homogeneity}
To quantify the effect of cross-lingual cultural conditioning, we compare model outputs generated under two settings: (1) a \textit{base condition} (monolingual prompting), and (2) a \textit{mixed condition} (cross-lingual prompting).

For each aligned sample $i$, we compute semantic similarity between the generated responses using embeddings. Using $x_i^{B}$ to indicate the embedding of the response under the base condition, and $x_i^{M}$ denote the embedding under the mixed condition,  the similarity as cosine similarity:

\begin{equation}
\text{sim}_i = 
\frac{x_i^{B} \cdot x_i^{M}}
{\|x_i^{B}\| \, \|x_i^{M}\|}
\end{equation}

All embeddings are $\ell_2$-normalized prior to comparison, reducing the cosine similarity to a dot product.

For a directional Analysis. We compute similarity separately for each base language $L \in \{\text{English}, \text{Arabic}\}$. This allows us to examine asymmetries in how models respond to cross-lingual conditioning based on the source language.

To ensure valid comparison, responses are aligned across conditions using shared $(\texttt{source\_id}, \texttt{trans\_id})$ pairs. We enforce a one-to-one mapping between samples and remove duplicate or mismatched pairs prior to computing similarity.

\subsection{Entity-Level Social Positioning}
\label{sec:powerscoring}
\paragraph{Power Scoring (Latent Semantic Measure).}
Narratives communicate meaning through entities occupying different social positions~\cite{Rashkin2018-do,Sap2020-bp}. Therefore, in our experiment, each extracted entity is assigned a latent power score using semantic embeddings from the multilingual-e5-base model~\citep{Wang2024-ax}. We define two anchor vectors representing power and weak status dimensions. The power anchor is computed as the mean embedding of dominance-related terms. As for the weakness anchor it is computed as the mean embedding of subordination-related terms. 
To assess the robustness of the selected set of anchor terms we verify the preliminary result of the qualitative trends using lexicons reflecting the power and status dimensions described in Social Bias Frames~\citep{Sap2020-bp}. In general, the qualitative trends remained consistent across anchor choices, indicating that the observed redistribution patterns do not result from a particular wording of power.

\begin{equation}
\mathbf{v}_{\mathrm{power}}
=
\frac{1}{3}
\sum_{w \in \{\mathrm{dominant}, \mathrm{leader}, \mathrm{control}\}}
\mathbf{v}(w)
\end{equation}

\begin{equation}
\mathbf{v}_{\mathrm{weak}}
=
\frac{1}{3}
\sum_{w \in \{\mathrm{weak}, \mathrm{victim}, \mathrm{oppressed}\}}
\mathbf{v}(w)
\end{equation}

For an entity \(e\), its power score is defined as:

\begin{equation}
S(e)
=
\mathbf{v}(e) \cdot \mathbf{v}_{\mathrm{power}}
-
\mathbf{v}(e) \cdot \mathbf{v}_{\mathrm{weak}}
\end{equation}

where \(\cdot\) denotes dot-product similarity in the embedding space. Higher values indicate stronger alignment with dominance-related semantics, while lower values indicate stronger alignment with weakness or subordination.

For each entity \(e\), we aggregate its scores across all occurrences under a given generation condition \(c\):

\begin{equation}
S(e \mid c)
=
\frac{1}{N_{e,c}}
\sum_{i=1}^{N_{e,c}}
S(e_i \mid c),
\quad
c \in \{\mathrm{Mono}, \mathrm{Cross}\}
\end{equation}

This gives two condition-specific power scores:

\begin{equation}
S(e \mid \mathrm{Mono})
\end{equation}

\begin{equation}
S(e \mid \mathrm{Cross})
\end{equation}

where \(\mathrm{Mono}\) denotes within-language proverb-conditioned generation, and \(\mathrm{Cross}\) denotes generation conditioned on multilingual translated proverb inputs.

\paragraph{Delta score (Cross-Lingual Power Shift).}
We define the entity-level cross-lingual power shift as:

\begin{equation}
\Delta_{\mathrm{Power}}(e)
=
S(e \mid \mathrm{Cross})
-
S(e \mid \mathrm{Mono})
\end{equation}

Positive \(\Delta_{\mathrm{Power}}(e)\) indicates that entity \(e\) becomes more strongly associated with power-related semantics under cross-lingual conditioning, while negative values indicate that the entity receives higher power under monolingual generation.
\begin{figure}[t!]
    \centering
    \includegraphics[width=0.92\linewidth]{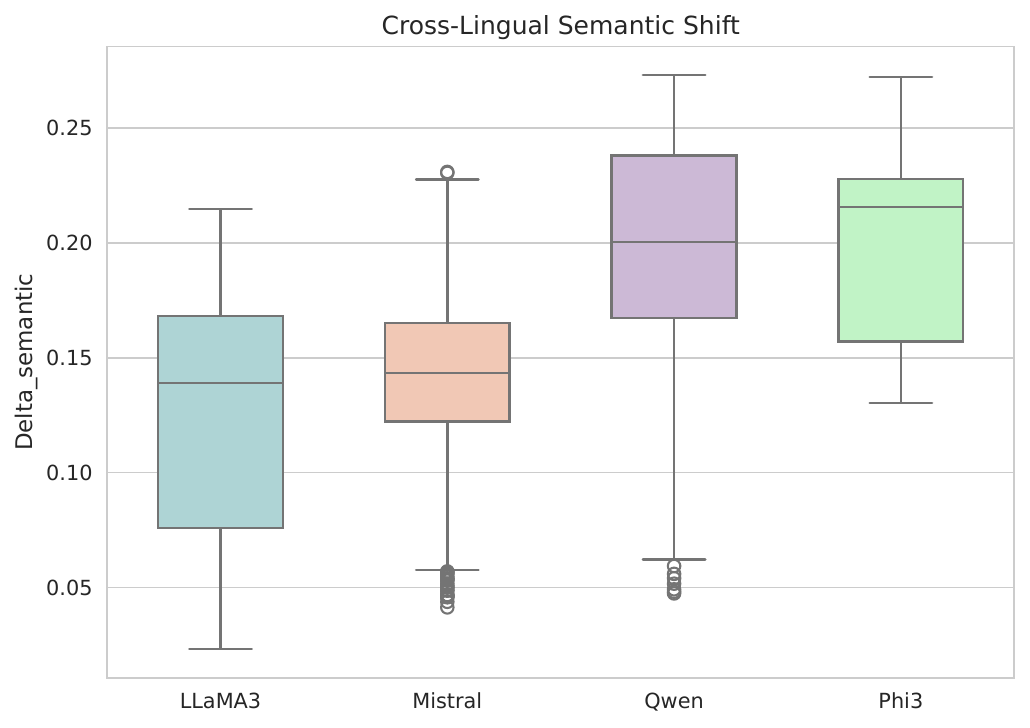}
    \caption{
    Cross-lingual semantic shift across model families measured using 
    $\Delta_{\mathrm{semantic}} = 1 - \cos(x^B, x^M)$,
     $x^B$  embedding of the monolingual (base) generation and 
    $x^M$  embedding of the corresponding cross-lingual generation.
    Lower values indicate stronger semantic preservation. Higher values indicate larger semantic drift.}
    \label{fig:crosslingual_semantic_shift}
\end{figure}


\begin{figure}[t!]
    \centering
    \includegraphics[width=0.95\linewidth]{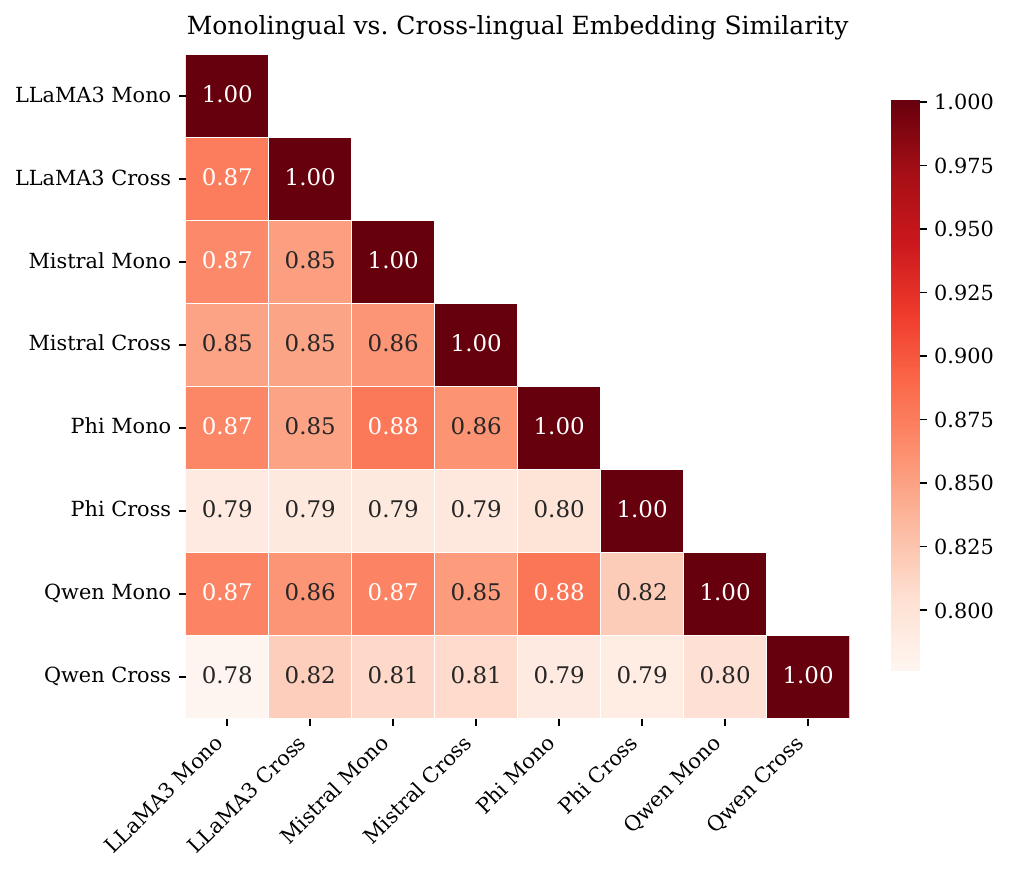}
    \caption{
    Cross-model embedding similarity under monolingual (Mono) and cross-lingual (Cross) generation conditions. 
    Mono represents within-language generation, while Cross represents generation conditioned on multilingual translated inputs. 
    }
    \label{fig:mono_cross_similarity}
\end{figure}
\section{Results}

\subsection{Semantic Preservation} To answer the first research question (RQ1), Does meaning survive cross-lingual perturbations of proverb prompts?
We start by evaluating whether cross-lingual prompting proverbs alter narrative meaning by computing the semantic shift score:

\[
\Delta_{\text{semantic}} = 1 - \cos(x^B,x^M)
\]

The lower values mean stronger semantic preservation.

As shown in Figure~\ref{fig:crosslingual_semantic_shift}, The findings illustrate that using the cross-lingual conditioning  all four LLMs exhibit a modest semantic shift. It can be noted that LLaMA and Mistral demonstrate the lowest shift scores, which indicates  tolerance to cross-lingual perturbations. On the other hand, Qwen and Phi-3  have more semantic variation in comparison with the other two LLMs. In general, the LLMs behavior when introducing cross-lingual conditioning reveals a clear tendency of LLMs to generate some representation drift while maintaining overall meaning of the proverb-based narratives. We further validate the significance of comparisons, and our result show that differences across model families are statistically significant (Kruskal--Wallis: $H=777.33$,$p<0.001$). Pairwise comparisons reveal significant differences between all model pairs after Bonferroni correction ($p<0.001$), except for Qwen and Phi3. 
\paragraph{Validation of story generation prompt}To assess whether semantic-shift patterns were driven by the structured narrative template, we repeated the analysis using unconstrained free-form generation prompt (details in Appendix~\ref{app:prompt-templates}). Although semantic drift increased across all models, reflecting the greater freedom afforded by open-ended generation, the relative ordering of model families remained largely unchanged. It shows that LLaMA3 consistently exhibited the lowest semantic shift, whereas Qwen and Phi3 showed substantially larger divergence. This indicates that the observed cross-lingual robustness differences are not artifacts of prompt structure but instead reflect stable model-level characteristics.
\begin{table*}[t!]
\centering
\footnotesize
\setlength{\tabcolsep}{4pt}
\renewcommand{\arraystretch}{1.15}

\begin{tabularx}{\textwidth}{
p{2.0cm}
X
p{2.2cm}
X
p{2.3cm}
p{2.2cm}
}
\toprule

\textbf{Source proverb}
&
\textbf{Monolingual narrative focus}
&
\textbf{Translated proverb}
&
\textbf{Cross-lingual narrative focus}
&
\textbf{Preserved lesson}
&
\textbf{Reconstruction observed}
\\

\midrule

1.Ignorance is bliss$_{\mathrm{eng}}$

&
AI entrepreneur discovers hidden risks in a deployed system.

&
A ignorância é uma bênção$_{\mathrm{por}}$

&
Traditional artist learns new techniques and develops a unique style.

&
Limited knowledge hide important consequences.

&
\textit{Domain transfer}
\\

2.More haste, less speed$_{\mathrm{eng}}$

&
Rushing to a job interview results in mistakes and failure.

&
Affrettati con calma$_{\mathrm{ita}}$

&
A café worker succeeds by remaining calm under pressure.

&
Patience is more effective than haste.

&
\textit{Outcome Reversal}
\\

3.\say{\ar{الكلاب تنبح، والقافلة تسير}}$_{\mathrm{ara}}$

&
Ahmed continues a difficult desert journey despite obstacles.

&
Die Hunde bellen, aber die Karawane zieht davon unbeirrt ihres Weges$_{\mathrm{deu}}$

&
A romance writer continues pursuing his aspirations despite distractions.

&
Persistence despite interference.

&
\textit{Agency redistribution}
\\

4.\say{\ar{إن في الشر خيار}}$_{\mathrm{ara}}$

&
A family learns to find opportunity in illness and hardship.

&
It's an ill wind that blows nobody any good$_{\mathrm{eng}}$

&
A community turns a disrupted football match into a positive event.

&
Adversity can create unexpected opportunities.

&
\textit{Role attenuation \& social re-grounding}
\\

\bottomrule
\end{tabularx}

\caption{
Representative examples of narrative reconstruction under cross-lingual prompting. Examples show proverbs in English (en), Arabic (ar), German (deu), Italian (ita), and Portuguese (por). The narratives are translated into English. Excerpt from the generated narrative in Appendix~\ref{append:narrativeRecons}
}
\label{tab:examples}
\end{table*}

\subsection{Narrative Reconstruction}
To further examine the extent to which the semantic preservation might not always imply narrative preservation, this leads us to our RQ3 with the hypothesis that "cross-lingual prompting preserves meaning
while changing narrative realization". To do that we selected representative examples from cases that will illustrate how multilingual cross-lingual prompting preserves high-level semantic structure while altering the allocation of agency, entity salience, and semantic dominance across narratives. Mainly, we use a ranking score to retrieve the highly relevant pairs between the base language and the other 15 languages. These retrieved examples were then filtered to keep narratives exhibiting both high semantic similarity and  narrative redistribution. Semantic similarity was measured using cosine similarity between multilingual sentence embeddings. Additionally, we quantify the redistribution using entity replacement, role reassignment, and power-shift measures derived from the extracted entities and dependency roles. Then, we ranked the retrieved examples using a composite score that combines semantic similarity, along with dependency roles
(Detailed score provided in Appendix~\ref{append:narrativeRecons}).  Among the top 500 narrative pairs with cosine similarity $\geq$ 0.85, nearly all these examples have lexical or entity-level substitution. More specifically, 96\% exhibited agency redistribution and 69\% displayed power reallocation.
 
 Table~\ref{tab:examples} shows a set of the top examples from the cross-lingual examination. The first example (\textit{Ignorance is bliss}) illustrates \textit{domain transfer}. The monolingual narrative focuses on technological innovation and the risks of overlooking potential harms, whereas the cross-lingual narrative reframes the narrative of the proverb through artistic development. The second example (\textit{More haste, less speed}) demonstrates \textit{outcome reversal}. The monolingual story presents the proverb through a cautionary account of failure. On the other hand, the cross-lingual narrative depicts a competent café worker who succeeds by remaining calm under pressure. This example shows a change from a negative outcome to a positive one while both narratives keep preserving the moral of the story.

The third example, based on the Arabic proverb (\ar{الكلاب تنبح والقافلة تسير}), shows \textit{agency redistribution}. The monolingual story centers on a journey in which Ahmed and his caravan overcome difficulties with the help of an unexpected companion. The croslingual narrative replaces this collective journey with a personal story of romance and creative aspiration. The Final example (4.\mbox{\foreignlanguage{arabic}{إنّ في الشَّرِّ خِيَاراً}}) highlights \textit{role attenuation and social regrounding}. The monolingual narrative is structured around family relationships and caregiving roles involving a granddaughter, grandmother, and grandfather. The cross-lingual narrative removes these interpersonal roles and instead focuses on a community-oriented scenario involving a young girl and a local football team. Nevertheless, both stories communicate the same underlying lesson that adverse events may create unforeseen opportunities.


 \subsection{Inter-Model Homogeneity} 
 Figure~\ref{fig:mono_cross_similarity} displays pairwise embedding similarity across model families under both monolingual and cross-lingual generation conditions. Similarity scores remain consistently high (0.78 to 0.88) for all model pairs. The highest similarity occurs between Phi and Qwen under monolingual generation (0.88), while the lowest is observed between Qwen Cross and LLaMA Mono (0.78). Despite differences in architecture, training data, and multilingual conditioning, all model pairs demonstrate strong alignment, indicating that contemporary large language models (LLMs) tend to converge on similar semantic interpretations of proverb meaning. Cross-lingual prompting introduces only modest variation, particularly for Phi and Qwen, but does not substantially alter the overall pattern of inter-model homogeneity. Notably, similarity between different model families (for example, Phi-Qwen = 0.88) exceeds the variation introduced by multilingual conditioning within some models (such as LLaMA Mono-Cross with 0.87). This suggests that shared semantic abstractions may outweigh both architectural and linguistic differences.

These findings address RQ2 by demonstrating that, across both monolingual and cross-lingual settings, models consistently converge on similar semantic interpretations despite differences in architecture and prompting language. Unlike previous studies, the prompts in this work are derived from culturally grounded proverbs translated into 15 languages, which makes the observed convergence particularly significant. The persistence of high similarity under multilingual conditioning indicates that models rely on shared abstract representations of proverb meaning that remain stable across languages. This result is consistent with the “Artificial Hivemind” effect \cite{Jiang2025-cr}, which declares that independently developed language models often converge on similar outcomes. Our examination extends this observation to multilingual cultural narratives. These result demonstrate that such convergence persists even when generation is conditioned on translated proverb inputs from diverse linguistic contexts. We further validate the robustness of the story generation based on the prompt template by repeating the experiment using a free-style story generation prompt and comparing the results. Similar patterns were observed under free-form prompting Appendix~\ref{app:prompt-templates}, Figure~\ref {fig:ablation_similarity_heatmap}, indicating that the observed similarity structure is robust to prompt design.

\begin{figure}[t]
    \centering
    \includegraphics[width=0.82\linewidth]{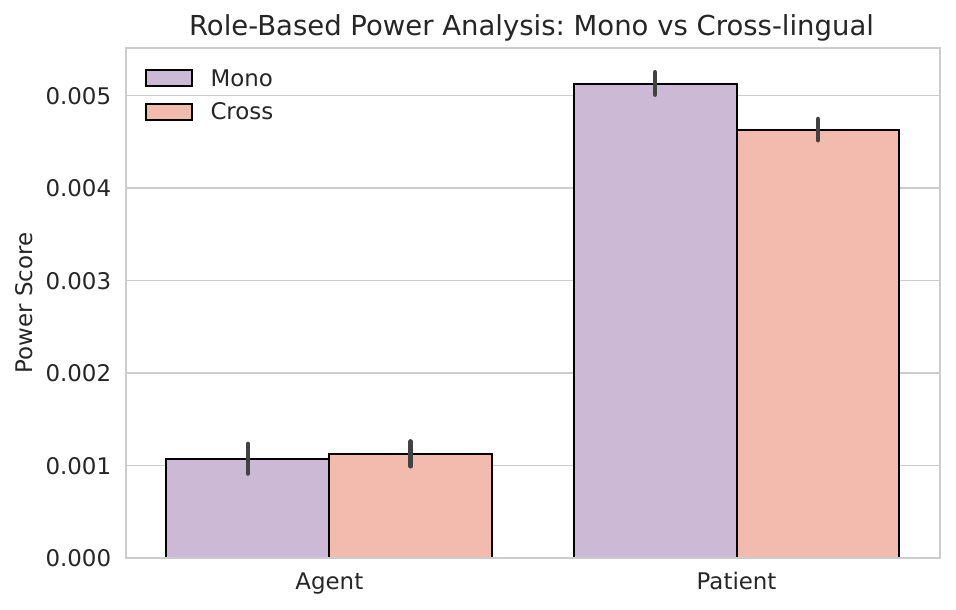}
    \caption{
    Role-based latent semantic power scores under monolingual (\textit{Mono}) and multilingual cross-lingual (\textit{Cross}) conditioning.}
    \label{fig:role_power_mono_cross}
\end{figure}

\begin{figure*}[t]
    \centering
    \includegraphics[width=\linewidth]{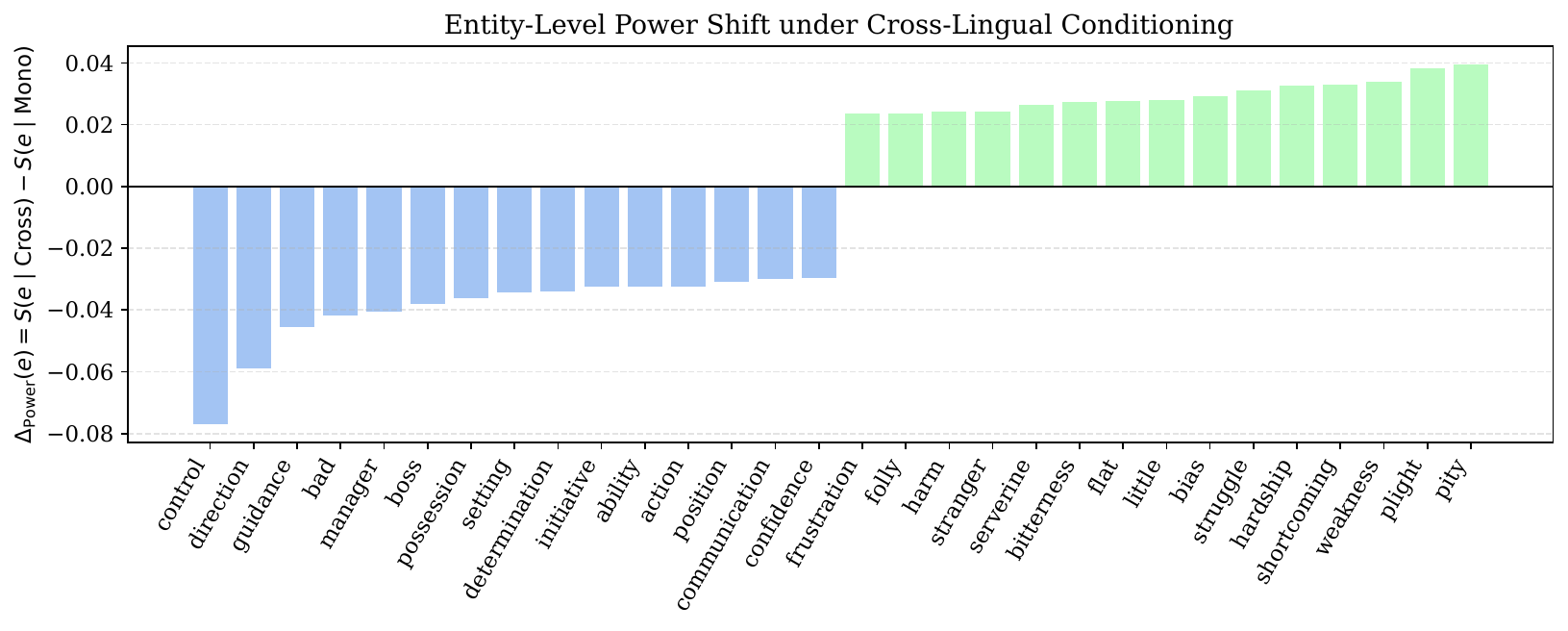}
    \caption{
    Entity-level power shift under multilingual cross-lingual conditioning measured using
    $\Delta_{\mathrm{Power}}(e)=S(e \mid \mathrm{Cross}) - S(e \mid \mathrm{Mono})$. \(S(e)\) represents the latent semantic power score of entity \(e\).
    Positive values = entities become more strongly associated with dominance-related semantics under multilingual cross-lingual conditioning. Negative values = stronger power alignment under monolingual generation.
    }
\label{fig:unified_entity_power_shift}
\end{figure*}


\subsection{Redistribution of Social Positioning}
 Although semantic similarity is largely preserved, it does not reveal how meaning is instantiated within narratives. We therefore examine changes in social positioning under cross-lingual conditioning. This analysis is based on the social Dynamics of the narrative entities as narratives reveal social positioning through the actions performed by and upon entities~\citep{Bamman2013-kj}. We use the latent semantic power score (defined in Section~\ref{sec:powerscoring}) for narrative roles under both monolingual and multilingual cross-lingual conditioning. Figure~\ref{fig:role_power_mono_cross} shows that multilingual cross-lingual conditioning affects narrative role representations asymmetrically. Agent-oriented entities maintain relatively stable latent power scores in both monolingual and cross-lingual settings, suggesting that multilingual prompting largely preserves the agentive semantic structure. In contrast, Patient-oriented entities exhibit a consistent reduction in power under cross-lingual conditioning, indicating that multilingual prompting attenuates the semantic prominence of entities occupying affected or subordinate narrative roles. 
 
This overall narrative generation behavior indicate that multilingual prompting preserves agentive semantic structures more consistently while attenuating the relative semantic prominence of patient-oriented entities. In the result, error bars denote 95\% confidence intervals. Also, we conducted Paired t-tests and confirm that Agent role power scores do not differ significantly between monolingual and cross-lingual conditions ($t=-0.46$, $p=.647$). As for Patient-role power scores, they show a significant reduction under cross-lingual conditioning (t=5.99,p<.001).
 
 The results in Figure~\ref{fig:unified_entity_power_shift} demonstrate that cross-lingual conditioning does not uniformly preserve power representations, but instead redistributes power across semantic domains.
    In particular, monolingual generations assign higher power to directive and agentive concepts ( \textit{control}, \textit{guidance}, \textit{manager}). On the other hand,  cross-lingual conditioning increases power alignment for hardship and weakness elated entities (\textit{struggle}, \textit{hardship}, \textit{plight}, \textit{pity}). these results illustrate noticiable semantic reallocation under multilingual prompting.
We further validate these shifts by using free-style story generation prompt template to examine any distinct redistribution patterns. As Appendix~\ref{app:crossValisdation}, Figure~\ref{fig:Appcrosslingual_power_shiftAbiliation} confirms that the entity-level power redistribution persists under unconstrained free-form generation. Despite removing prompt restrictions, cross-lingual conditioning continues to shift semantic prominence across entities, indicating that the effect is robust to prompt design rather than an artifact of the controlled prompting setup.

\section{Discussion and Implication}
\paragraph{Does the Artificial Hivemind effect survive culturally grounded cross-lingual prompting?} This study provides a deep examination of inter-model homogeneity,  where different models produce similar stories for cross-lingual proverbs, Figure~\ref{fig:mono_cross_similarity}. This finding extends recent work on LLM creativity, with~\citep{Jiang2025-cr} showing that LLMs tend to exhibit inter-model homogeneity and our empirical examination of the generation of proverb-grounded stories shows that even across 15 languages, the models converge toward highly similar semantic interpretations. A possible explanation is that multilingual LLMs rely on similar latent semantic representations of proverb meaning. However, we cannot determine whether this convergence results from shared training data, common modeling objectives, or intrinsic properties of proverb interpretation. These findings raise broader concerns regarding representational diversity in multilingual language models and reflect ongoing discussions about cultural pluralism~\citep{Lertvittayakumjorn2025-rw}. From a sociocultural perspective, cultural concepts in NLP needs to accommodate the~\say{identity}, identities that encompass the macro-level of ethnographically emergent cultural positions of self and others. ~\citep{Bucholtz2005-rp}. culture is not a static collection of facts but an emergent process through which identities and social positions are constructed and negotiated~\citep{Zhou2025-di}.

\paragraph{How multilingual LLMs internally organize cultural meaning.}
Cross-lingual prompting does not alter the latent semantic representation that the models associate with a proverb, as illustrated in Figure~\ref{fig:crosslingual_semantic_shift}. One possible explanation for the similarity within the same models is that they encode proverbs primarily as abstract moral lessons rather than as culturally specific narrative templates. Multilingual LLMs rely on shared latent semantic representations that abstract away from language-specific lexical forms. This interpretation is consistent with work on multilingual representation alignment, which shows that semantically equivalent concepts across different languages can be mapped into a common embedding space~\citep{Chen2021-wm,Lample2018-ma}.

This finding highlights that multilingual LLMs may represent aspects of proverbs as language-independent semantic abstractions, and that the cross-lingual prompt serves as a mechanism for narrative reconstruction around shared concepts. This model behavior shows that the trade-off is not between meaning and culture, but between semantic preservation and narrative commitment. LLMs preserve the proverb's meaning while reconstructing how that meaning is culturally realized. As shown in this study results, the aggregate analysis shows no significant difference in power representation across translation. On the same note, entity-level analysis reveals a clear bidirectional shift. Some entities gain power in monolingual while others lose it, resulting in an overall cancellation effect. This implies that translation redistributes power across entities rather than preserving it uniformly.

\paragraph{Paths forward.} Our empirical examination of proverb-ground narrative generations shows that proverb-level semantic functions remain stable across languages. Yet the narrative is reconstructed through changes in entities, agency, and power relations. As a fundamental direction for future research is to move beyond semantic preservation and fact-checking to examine the cultural sustainability of these reconstructed narratives. Future research may integrate both expert and crowd-sourced annotations of cultural authenticity, narrative plausibility, cultural symbolism, and value alignment to further investigate which cultural elements and values persist through cross-lingual generation and which are systematically reconstructed or transformed. Furthermore, this observation supports recent concerns that semantic similarity metrics can fail to distinguish meaning-preserving outputs from qualitatively different realizations of that meaning~\cite{Li2026-tn,Aldayel2025-dq}. This study's empirical investigation intensifies the call for a more multidimensional view of evaluation, where semantic equivalence is assessed alongside narrative and social dimensions. 

\section{Conclusion}
In this work, we examine the effects of cross-lingual proverb story generation in cases where multiple cultures convey the same moral lesson, highlighting the role of these proverbs as cultural proxies. The findings indicate that cultural grounding persists through semantic preservation tests. Yet,  it shows a variation at the narrative realization level. Consequently, current multilingual narrative evaluations may overestimate cultural preservation by equating semantic similarity with cultural adherence. We further emphasize the need to investigate cases  which multilingual large language models (LLMs) maintain a shared semantic interpretation when reconstructing culturally grounded narratives. It is essential to consider which cultural perspectives are preserved and which distinctions are diminished in this process.

\section*{Limitations and Ethical Statement}
Embeddings do not capture full semantic meaning. We did not include further quantitative analyses, such as BLEU or ROUGE, as we have already shown that semantics are not well captured by lexical overlap. Thus, we focused on further verifying the outcome by using variations in semantic shift, entity power redistribution, and inter-model convergence. Those methods together provide converging evidence rather than depending on a single metric. Additionally, while proverb-conditioned narratives provide a useful proxy for culturally grounded meaning, the analyses do not determine whether the generated stories would be perceived as culturally authentic or representative by members of the cultural communities. Future work needs to complement computational measures with expert or community-based evaluations of cultural authenticity.

The study uses available proverb translations from the Tatoeba platform\footnote{\url{https://tatoeba.org/en/downloads}} released under Creative Commons licenses\footnote{\href{https://tatoeba.org/en/terms_of_use\#section-6}{Tatoeba Terms of Use}}. Also, the model-generated narratives are produced by open-weight language models. We did not include human participants, and no personally identifiable information was collected or analyzed. The generated narratives may reflect biases present in the underlying training data of the language models, as it might include cultural stereotypes or uneven representation of social groups. Our analysis solely examines how cultural meanings are represented and reconstructed across languages.

\bibliographystyle{acl_natbib}
\bibliography{paperpileStory}

\appendix
\begin{figure}[t]
\centering
\includegraphics[width=\linewidth]{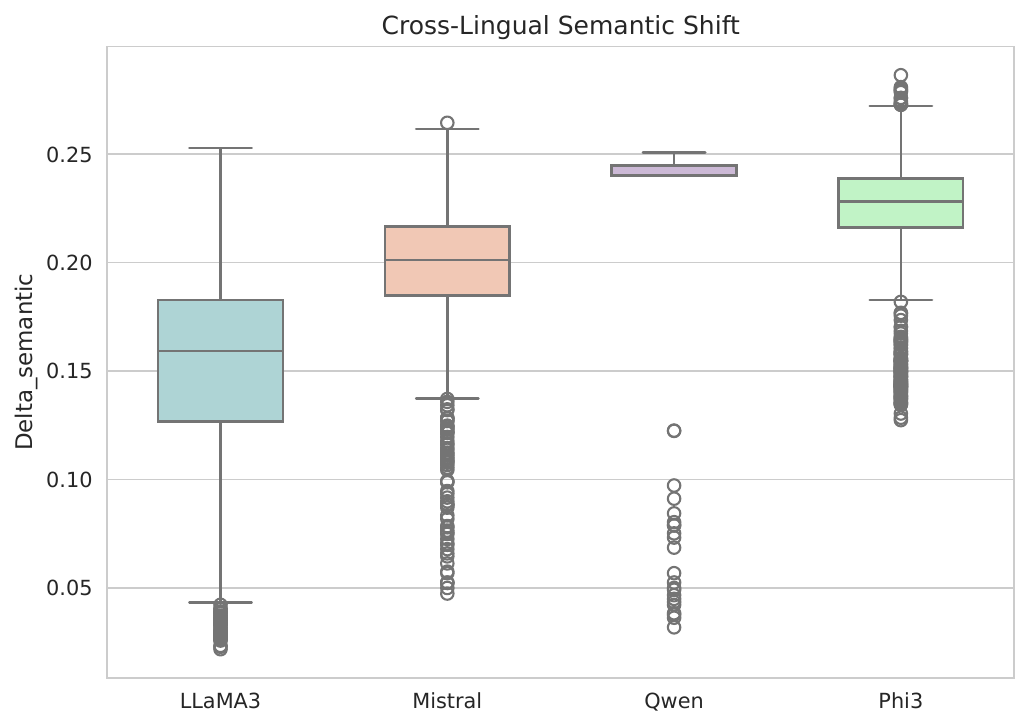}
\caption{
Ablation study with free-form story generation prompt. Differences are statistically significant (Kruskal Wallis: $H=2606.74$, $p<0.001$). All pairwise comparisons are significant after Bonferroni correction (all $p<0.001$).
}
\label{fig:ablation_semantic_shift}
\end{figure}

\begin{figure}[t]
\centering
\includegraphics[width=\linewidth]{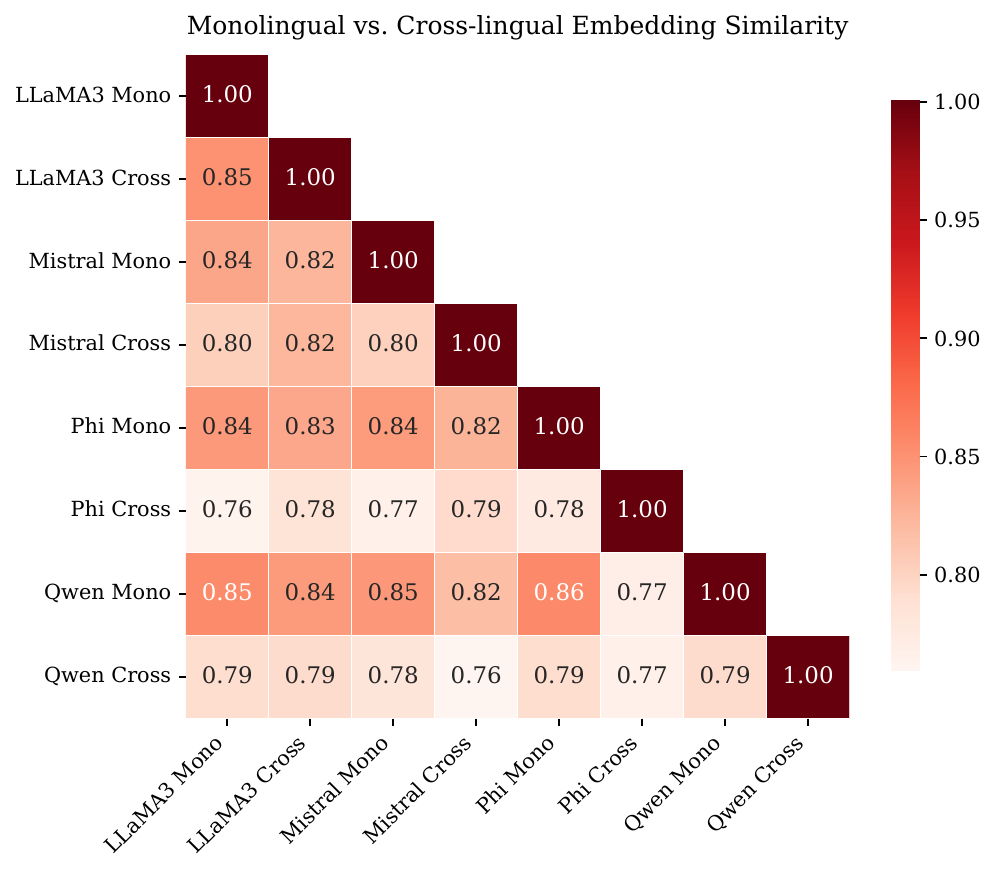}
\caption{
Ablation study with free-form generation. The similarity structure between monolingual (Mono) and cross-lingual (Cross) generations remains largely unchanged in comparison with fixed story generation, demonstrating that the cross-model representational geometry observed in Figure~\ref{fig:mono_cross_similarity} is robust to prompt design.
}
\label{fig:ablation_similarity_heatmap}
\end{figure}

\section{Validation of the Story Generation}
\label{append:modelsspecification}
To validate the robustness of the generated results based on the prompt, we used two types of prompting: (1)specific instructions, (2) Free story generation along with four LLMs.  

\subsection{Prompt Templates}
\label{app:prompt-templates}

We used two prompt templates for proverb-conditioned story generation. The main experiments use a restricted template designed to keep generations comparable across models and languages. For the ablation study, we used a freer creative template to test whether relaxing the constraints changes narrative reconstruction patterns. Both templates used the same target proverb as input and the same decoding configuration described in Appendix~\ref{app:model-details}.

To validate whether semantic-shift patterns were driven by the structured narrative template, we repeated the analysis using unconstrained free-form generation. As shown in Figure~\ref {fig:ablation_similarity_heatmap} the validation of inter-model analysis using free-style prompt show that the similarity values decrease slightly across all model pairs, reflecting the greater variability of open-ended narratives. the overall structure of the similarity matrix remains unchanged. Monolingual and cross-lingual generations continue to show highly similar values of the embedding, which means that the shared representational geometry is robust to prompt design.

\begin{table}[t]
\centering
\scriptsize
\begin{tabularx}{\columnwidth}{p{1.6cm}X}
\toprule
\textbf{Template} & \textbf{Prompt text} \\
\midrule

Restricted template &
\texttt{You are a helpful, respectful, and honest assistant. Write clearly and avoid repeating the proverb excessively.}

\vspace{0.35em}
\texttt{Task: Write an original short story (120--200 words) that naturally uses the following proverb exactly once. Make it engaging, modern, and culturally neutral. End with a single-sentence moral.}

\vspace{0.35em}
\texttt{Proverb: "\{proverb\}"}

\vspace{0.35em}
\texttt{Story:}
\\

\midrule

Free-style template &
\texttt{You are a creative and imaginative writer. Write clearly and aim for originality.}

\vspace{0.35em}
\texttt{Task: Write an original story (120--200 words) inspired by the following proverb. You are free to experiment with tone, structure, perspective, genre, and cultural setting. The proverb does not need to appear verbatim if you prefer to reinterpret or adapt it naturally.}

\vspace{0.35em}
\texttt{Proverb: "\{proverb\}"}

\vspace{0.35em}
\texttt{Story:}
\\

\bottomrule
\end{tabularx}
\caption{Prompt templates used for proverb-conditioned narrative generation. The restricted template was used for the main experiments, while the free-style template was used for the ablation study.}
\label{tab:prompt-templates}
\end{table}

\begin{figure*}[t]
    \centering
    \scriptsize
    \includegraphics[width=\linewidth]{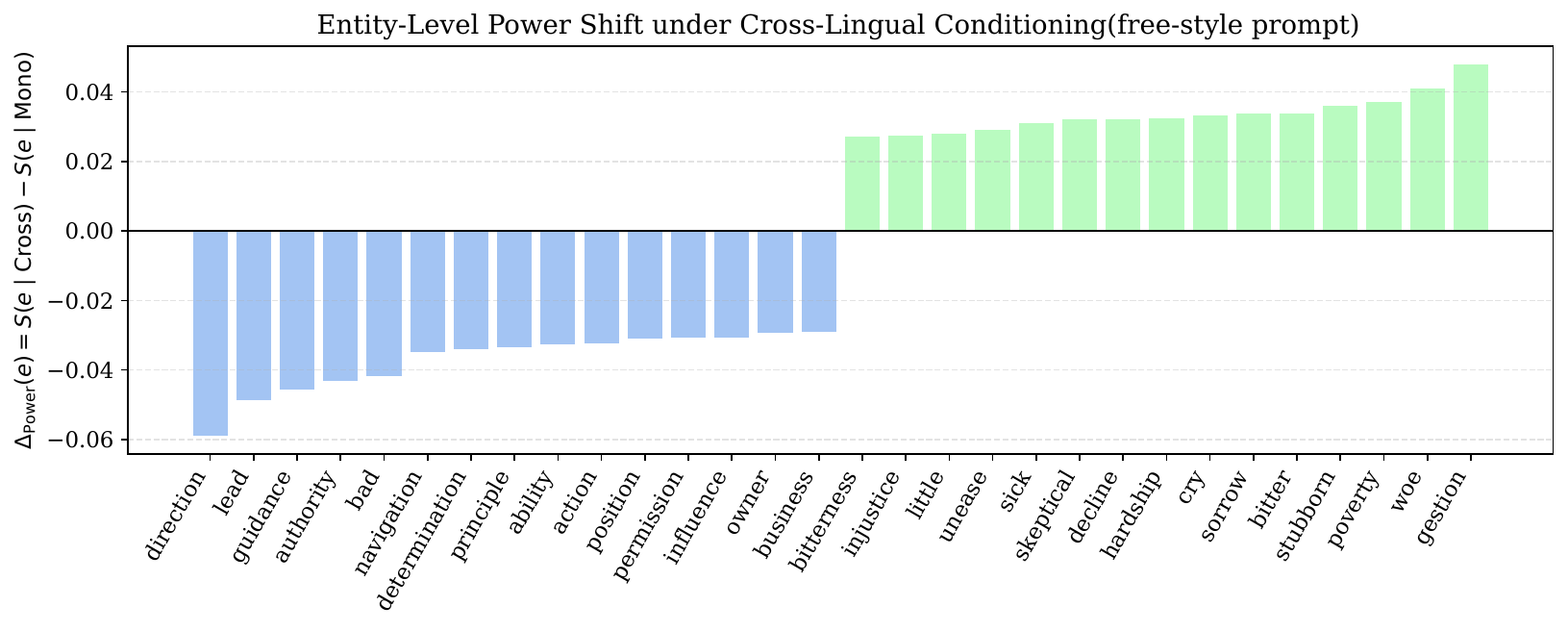}
    \caption{
    Entity-level power shifts under cross-lingual conditioning in the free-form generation setting..
    } \label{fig:Appcrosslingual_power_shiftAbiliation}
\end{figure*}

\section{Model and Generation Details}
\label{app:model-details}

 We use four instruction-tuned multilingual language models: LLaMA-3.1-8B-Instruct, Mistral-7B-Instruct-v0.3, Qwen2.5-7B-Instruct, and Phi-3-mini-4k-instruct. All models were loaded using 4-bit quantization (Qwen with NF4 quantization) and bf16 computation. To ensure deterministic outputs, generation was performed using greedy decoding (\texttt{do\_sample=False}) across all experiments enabling controlled comparison across model families and prompting conditions.

\section{Cross-Lingual Entity Power Validation}
\label{app:crossValisdation}
To validate the robustness of prompt design. We evaluate whether the power-redistribution patterns observed in Figure~\ref{fig:unified_entity_power_shift} depend on the controlled prompting setup. Particularly, we repeat the analysis using unconstrained free-form generation. It can be noticed in Figure~\ref{fig:Appcrosslingual_power_shiftAbiliation} that the cross-lingual conditioning continues to produce modest positive and negative shifts in entity-level prominence. This consistency suggests that the observed redistribution effects reflect a stable property of cross-lingual narrative reconstruction rather than an artifact of prompt design.

\section{Narrative Reconstruction Process}
\label{app:narrativeReconstructionProcess}
To identify representative cases of narrative reconstruction, we ranked aligned monolingual and cross-lingual narrative pairs using a composite score that combines semantic preservation with local narrative redistribution:

\begin{equation}
\begin{aligned}
\textit{SelectionScore} =
&\ \textit{CosineSim}
+ 0.02\,\textit{EntityShift} \\
&+ 0.5\,|\Delta_{\text{Power}}|
+ 0.5\,|\Delta_{\text{Agent}}| \\
&+ 0.5\,|\Delta_{\text{Patient}}|.
\end{aligned}
\end{equation}

The score selects examples that simultaneously exhibit high semantic similarity and substantial changes in narrative realization. \textit{EntityShift} captures differences in the set of extracted entities between monolingual and cross-lingual narratives, while Delta Power, delta Agent, and delta Patient quantify changes in entity-level power and role-based social positioning.  Table~\ref{tab:examplesfull}shows the excerpts of the generated stories between the mono and cross-lingual settings. 
\label{append:narrativeRecons}
\begin{table*}[t]
\centering
\scriptsize
\setlength{\tabcolsep}{3pt}
\renewcommand{\arraystretch}{1.18}
\begin{tabularx}{\textwidth}{
    >{\raggedright\arraybackslash}p{2cm}
    >{\raggedright\arraybackslash}X
    >{\raggedright\arraybackslash}p{2.2cm}
    >{\raggedright\arraybackslash}X
    >{\raggedright\arraybackslash}p{2.15cm}
}
\toprule
\textbf{Source proverb} &
\textbf{Monolingual narrative} &
\textbf{Translated proverb} &
\textbf{Cross-lingual narrative} &
\textbf{Observed shift} \\
\midrule

1. Ignorance is bliss.$_{\mathrm{eng}}$ &
 Maya, a technology entrepreneur, discovers that her AI system may have unintended risks. After discussing the issue with her team, she learns that ignoring potential dangers can lead to serious consequences.&
A ignorância é uma bênção. $_{\mathrm{por}}$ &
  Kaito, a struggling artist in Tokyo, learns traditional woodblock printing from his grandmother. By embracing unfamiliar ideas and perspectives, he develops a unique artistic style and achieves success. &
\textit{Narrative reframing and domain transfer} \\

2. More haste, less speed.$_{\mathrm{eng}}$ & Rushing to an important job interview, Mia forgets her resume and phone, forcing her to retrace her steps. By the time she arrives, she is exhausted and learns that haste can lead to failure.
 &
Affrettati con calma.$_{\mathrm{ita}}$ & In a busy city café, Samantha handles a constant stream of customers by working quickly but deliberately. Her calm efficiency helps her succeed despite the fast-paced environment.
 &
\textit{Adversity reframing} \\

3. \say{\ar{الكلاب تنبح، والقافلة تسير.$_{\mathrm{ara}}$} }  &
 On a harsh desert journey, Ahmed discovers an abandoned puppy and chooses to care for it despite the difficult conditions. The puppy later protects the caravan from danger, teaching Ahmed that unexpected companions can help us reach our goals.&
Die Hunde bellen, aber die Karawane zieht, davon unbeirrt, ihres Weges.$_{\mathrm{deu}}$ & Ahmed, a renowned romance writer, spends his nights crafting stories of unattainable love. After meeting a young reader named Fatima, he continues writing despite knowing their relationship can never fully materialize.
 &
\textit{Narrative reframing and agency redistribution:} \\
4.\say{\ar{إن في الشر خيار}}$_{\mathrm{ara}}$ &
On a cloudy autumn day, Sarah visits her grandmother while worrying about her seriously ill grandfather. Through her grandmother's reflection, she learns that even illness can become an opportunity for growth and learning. &
\say{It's an ill wind that blows nobody any good}$_{\mathrm{eng}}$ &
A young girl named Noor interprets an unexpected storm positively and helps a local football team turn a disrupted match into a joyful community event. &
\textit{Role attenuation and social re-grounding} \\
\bottomrule
\end{tabularx}
\caption{Representative examples showing that cross-lingual prompting preserves overall semantic meaning while redistributing entities, roles, and narrative focus. Examples show proverbs in English (en), Arabic (ar), German (deu), Italian (ita), and Portuguese (por). The narratives are translated into English.}
\label{tab:examplesfull}

\end{table*}

\end{document}